\documentclass[letterpaper]{article} % DO NOT CHANGE THIS
\pdfoutput=1
\usepackage{aaai23}  % DO NOT CHANGE THIS
\usepackage{times}  % DO NOT CHANGE THIS
\usepackage{helvet}  % DO NOT CHANGE THIS
\usepackage{courier}  % DO NOT CHANGE THIS
\usepackage[hyphens]{url}  % DO NOT CHANGE THIS
\usepackage{graphicx} % DO NOT CHANGE THIS
\usepackage{natbib}  % DO NOT CHANGE THIS AND DO NOT ADD ANY OPTIONS TO IT
\usepackage{caption} % DO NOT CHANGE THIS AND DO NOT ADD ANY OPTIONS TO IT
\usepackage{algorithm}
\usepackage{algorithmic}

\usepackage{booktabs}       % professional-quality tables
\usepackage{amsfonts}       % blackboard math symbols
\usepackage{nicefrac}       % compact symbols for 1/2, etc.
\usepackage{microtype}      % microtypography
\usepackage{xcolor}         % colors

\usepackage{verbatim}
\usepackage{amsmath}
\usepackage{graphicx,psfrag}
\usepackage{enumerate}

\usepackage{newfloat}
\usepackage{listings}
\DeclareCaptionStyle{ruled}{labelfont=normalfont,labelsep=colon,strut=off} % DO NOT CHANGE THIS
\floatstyle{ruled}
\newfloat{listing}{tb}{lst}{}
\floatname{listing}{Listing}
\title{Novelty Detection for Election Fraud: A Case Study with Agent-Based Simulation Data}

\author {
    Khurram Yamin\equalcontrib,\textsuperscript{\rm 1}
    Nima Jadali\equalcontrib,\textsuperscript{\rm 2}
    Dima Nazzal,\textsuperscript{\rm 1}
    Yao Xie \textsuperscript{\rm 1}
}
\affiliations {
    \textsuperscript{\rm 1} Georgia Tech, Department of Industrial and Systems Engineering \\
    \textsuperscript{\rm 2} Georgia Tech, College of Computing \\
    kyamin3@gatech.edu, njadali3@gatech.edu, dima.nazzal@gatech.edu, yao.xie@isye.gatech.edu
}
\nocopyright

\begin{document}

\maketitle
\begin{abstract}
In this paper, we propose a robust election simulation model and independently developed election anomaly detection algorithm that demonstrates the simulation's utility. The simulation generates artificial elections with similar properties and trends as elections from the real world, while giving users control and knowledge over all the important components of the elections. We generate a clean election results dataset without fraud as well as datasets with varying degrees of fraud. We then measure how well the algorithm is able to successfully detect the level of fraud present. The algorithm determines how similar actual election results are as compared to the predicted results from polling and a regression model of other regions that have similar demographics. We use k-means to partition electoral regions into clusters such that demographic homogeneity is maximized among clusters. We then use a novelty detection algorithm implemented as a one-class Support Vector Machine where the clean data is provided in the form of polling predictions and regression predictions. The regression predictions are built from the actual data in such a way that the data supervises itself. We show both the effectiveness of the simulation technique and the machine learning model in its success in identifying fraudulent regions. 
\end{abstract}

\section{Introduction}
The aftermaths of elections have been fraught with fraud allegations for as long as elections have existed. In the United States, this has become a contentious issue for debate among the general public and policymakers alike. It has become especially relevant after the 2020 Presidential Election in which President Trump alleged that the Democrats committed wide-scale voter fraud \cite{yen_2021}. Given this, the need for a robust statistical methodology to detect potential fraud is evidently clear.

A problem faced when creating and testing methodologies to detect potential fraud is access to truly clean and fraudulent datasets. Our simulation addresses this by generating elections where the amount of injected fraud is known completely, making the evaluation and comparison of models possible and fair. The simulation generates a dataset based on an artificial election with similar properties and trends as a real-world election. Variables such as the number of electoral regions and the level of known fraud can be controlled unlike in actual elections making evaluating and comparing anomaly detection performance possible. The simulated election has greater control over variability as a multitude of possible elections with various sizes, constitutions, and random noise can be created. To generate a simulated election dataset, individuals which populate different electoral regions are instantiated. Each individual votes for one of two candidates based on each individual's attributes, residing electoral region, and injected random noise. A poll is cast on a small percent of the simulated population and will be used by anomaly detection algorithms. The final step is to inject a controlled amount of fraud into the election, potentially changing the election results in favor of one of the candidates. The polling data, election results, and injected fraud are saved and can be provided to anomaly detection algorithms to analyze performance.

It may not be possible for an algorithm that does not operate on an individual ballot level to detect definitive fraud given the highly variable nature of voting. As such, the algorithm we developed takes into account actual results and polling data to provide suggestions of potentially fraudulent voting regions. Its creation draws itself from two of our assumptions. First, we presume that, in general, regions with similar demographics that exist in close proximity to each other should have similar election results. Secondly, we presume that polling has some degree of value in predicting the election results in the location in which the algorithm is being run upon. In the United States, for example, there is a 2.9 $\%$ polling error on average in national elections when the polls are aggregated according to a trustworthiness index invented by the popular election analytics site 538 \cite{geoffreyvs_2021}.

Our paper is organized in the following order. We first review the relevant published literature on election simulation as well as on the use of statistical methods for elections. We then detail the methodology behind the agent-based simulation. This can be further subdivided into a simulation overview, election generation, polling, and fraud injection. We then detail the methodology of the machine learning model used to detect anomalies. This can be subdivided into the regression, primary clustering, and novelty detection steps. Lastly, we apply our methodology to simulated election data and analyze the results in comparison to several state-of-the-art baseline models. It is important to note that separate researchers on our team created the simulation and the machine learning model, and neither had access to each other's source code. We did this to maintain the credibility of the model as well as its unbiased nature.
\raggedbottom

\section{Literature Review}
Partially simulated data has been commonplace in studies involving elections and election fraud. One study used simulated data that would sample electoral results from a distribution based on a real-world election \cite{ShikanoMack+2011+719+732}. This approach has the benefit of making very little assumptions and generating data that reliably mimics the real world. However, this approach assumes that no fraud existed in the sampled electoral results. In addition, the lowest level of manipulation and data aggregation is on the precinct level which makes injecting fraud on the individual voter level impossible. The lowest level of granularity of the data produced by the simulation makes a significant difference when it comes to comparing and evaluating different methodologies. At the precinct level, fraud can only be injected by changing the results of elections by a percentage. At the individual level, different types of more detailed and controlled fraud can be injected such as swapping, deleting, or adding the votes of individuals with certain attributes or voting patterns within a portion of the precincts.

Another popular study testing Benford's Law and its relevance to election fraud detection, using a simulated dataset, took a more meticulous approach by having their simulation run on the individual level \cite{deckert_myagkov_ordeshook_2011}; however, this simulation generates homogeneous electoral regions. While homogeneous regions and distributions are easy to generate, they diverge from reality and trivialize the detection of injected anomalies. We utilize a similar approach involving a simulation at the individual level; however, our simulation differs by generating diverse electoral regions with demographic distributions and trends more similar to real life.

Recently, there have been several attempts to use machine learning for detecting fraud in elections. \cite{levin_pomares_alvarez_2016} used several popular methods and discovered that k-means shows efficacy in detecting clusters that were synthetically created in election data. We utilize this discovery in our first clustering step. \cite{cantu_2019} utilizes computer vision to detect changes in ballots in terms of physical alterations of ballots. However, such an approach can only work to detect one specific type of fraud while we seek to design a more holistic model. A work by \cite{10.1145/3530190.3534799} details the use of a machine learning method to attempt to detect generalized fraud. The methodology uses an unsupervised two-step clustering approach to identify outliers that don't match the election results of similar electoral regions. It shows efficacy in detecting very little fraud in an election that was confirmed to have little fraud through a full recount and investigation and detected a high amount of fraud in an election that was known to be largely fraudulent. We believe that we can build upon this framework by making it supervised and taking into account polling data which makes it more sensitive to detecting cases of mass fraud as polling is theoretically independent from the election governance. \cite{zhang_alvarez_levin_2019} details another attempt at using machine learning for election fraud detection. They attempt to simulate, using regression, a clean dataset and inject fraud by subtracting or adding votes from the vote totals. Then they use this to train a random-forest model, which is applied to a real-world case. We compare our methodology to both \cite{10.1145/3530190.3534799} and \cite{zhang_alvarez_levin_2019} as baselines.   

\section{Methodology}
\subsection{Election Generation}

The minimum information needed to generate the population is the number of electoral regions, population size, a list of attributes with their likelihoods, and a target election result. We based our population attributes on the 2000 US census \cite{census:2000}; however, any variables and their probability distribution could be used. The probability distributions are used to set the likelihoods for certain attributes possessed by individuals when they are being randomly instantiated. For example, we could describe the average income of a population using a Gaussian distribution with a specified mean and standard deviation, so when individuals are being instantiated their income is sampled from this distribution. As shown in Table 1, we see that our simulation generates a population similar to the real-world distributions it is based on. We could also use random sampling to assign attributes to individuals based on a premade population.

\begin{table}[h!]
\centering
\begin{tabular}{llllllll}
              \midrule
              & & Income Levels \\ 
              & Lower Class & Middle Class & Upper Class \\
              \midrule
             Census & 37.8\% & 44\% & 18.2\% \\
             Sim. & 39\% & 41.2\% & 19.8\%\\
\end{tabular}
\caption{This table shows the similarity in distributions of income for income groups between the Census, the 2000 US census, and Sim., a run of our simulation. The income groups are as followed: lower class is $< \$49,999$, middle class is $\$50,000 - \$149,999$ and upper class is $> \$150,000$. This specific simulation run has about 250 precincts with a total of 500,000 randomly generated individuals and used the 2000 US census distributions as a baseline for attribute likelihoods in its population.}
\end{table}

To create variety in the population makeup of different electoral regions, each electoral region has a portion of its population redistributed. The redistribution works by sampling a selected portion of an electoral region's population and computing a desirability score for each electoral region based on the individual's attributes and each electoral region's desirability probabilities associated with each attribute. The higher the desirability score, the more likely an individual is to relocate to the associated electoral region. If one electoral region has a higher desirability probability for an attribute $A$, then individuals with this attribute will be likely to redistribute to this electoral region. As a countermeasure against too much redistribution generating an unbalanced population, electoral regions with too large of a population can no longer gain individuals.

For voting, a neural network with randomly instantiated weights takes in a list of an individual's attributes and outputs a score used to determine an individual's vote. This neural network has one hidden layer 2 times the size of the input. The hidden layer has a random dropout chance where it zeros out the layer's output weights to add randomness to the voting score. The sum of the hidden layer is used to determine an individual's voting score. Then a threshold is used to split all the voting scores of the entire population between the two candidates. For example, if the median of all voting scores is chosen, then the threshold would divide the votes evenly between the two candidates. By basing our overall threshold on the sum of voting scores for the entire population, we can control the overall popular vote results while maintaining electoral region-level variations in the percent votes candidate received. 

An important assumption with this approach to generating votes is that individuals' votes are heavily influenced by certain key attributes that can describe them. These key attributes, such as income or political affiliation, can then be used to predict who an individual votes for. If two individuals have the same exact attributes, then the neural network and thresholding would determine that their votes would be for the same candidate if not for the dropout and additional noise added to account for other unaccounted influences on an individual's vote. Drop-out noise zeros out the weights of the neural network at random when it is predicting which candidate an individual votes for. The intuition of this is that not all values and attributes hold the same weight for individuals within a population; therefore, we need variability in the neural network weights on the individual level. Another type of randomness is noise injected into the voting scores to change who an individual votes for. The noise is sampled from a small, uniform distribution, so that the closer an individual's voting score is to the threshold between voting for candidates A and B, the more likely they are to have their vote flipped.

% Desired divisions are just how many precincts or sections there should be. Each division will have its election results determined differently from other divisions. Divisions will also have their own unique population makeup where features like the average income or racial distribution will differ from precinct to precinct.

We also split the populations into two categories, voters who cast their votes in person and those who mail in their votes. Whether an individual votes mail-in or votes in person is an attribute inherent to each individual and influenced by their other attributes. This assignment is similar to the redistribution where we sample a portion of the population, and from that portion we calculate a mail-in preference score. The higher the score, the more likely an individual is to vote by mail.

\begin{figure}[h!]
    \centering
    \begin{tabular}{cc}
    \includegraphics[width = .35\textwidth]{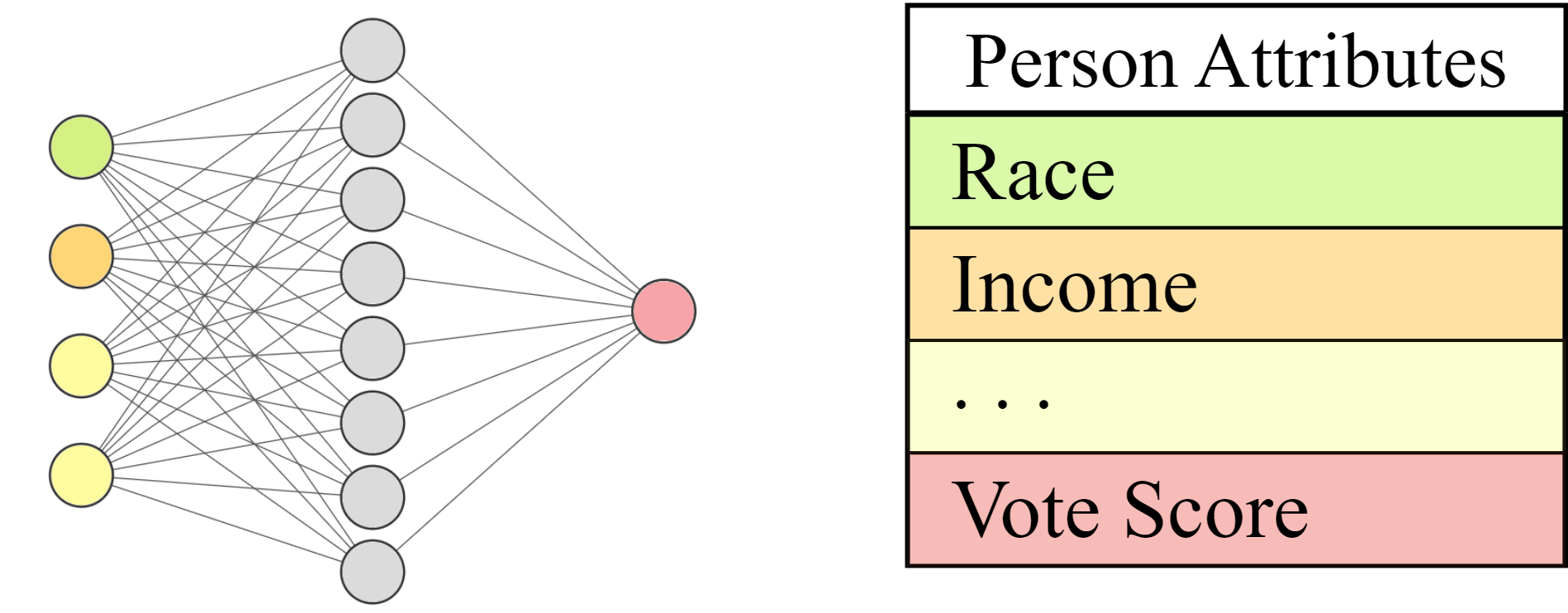}\\
    \end{tabular}
    \caption{The Neural Network takes in the attributes of an individual and outputs a voting score which can be used to compute an individual's vote}
\end{figure}

% \subsubsection{Assumptions}
% People’s votes can be simulated using a limited number of attributes that carry various weights in determining who someone votes for. Randomness is added to reflect that chance of someone on a whim changing their votes to opposing candidates or being influenced by some unique unaccounted event outside the attributes used. Assumption on independence are made with the polling for the generation of covariance matrices.

% \subsubsection{Simulation Architecture}
% The neural network that predicts individuals' votes based on their attributes is called VP-NN. The VP-NN architecture is a simple two layer neural network with a hidden layer 2 times the size of the input. The hidden layer is connected to a dropout layer as well that randomly zeros out the layer's output weights to add some randomness. The neural network then outputs the sum of all the nodes in the second layer. The median of all the sums can be used as a threshold to divide all the votes to the targeted split.

% \subsubsection{Randomness}
% Talked about above, but add a more detailed explanation of everything

\subsection{Polling}
The polling represents unbiased polling done on elections. Once the population generation and voting is complete, we sample a small portion of the population to see both their attributes and votes. We polled $5\%$ of our population in our experiments. Random noise is also injected into the poll results to replicate the same amount of error polling in the real world has. We aimed for an average error of 2.9\% as that is approximately the error achieved by the average polls of United States national elections \cite{geoffreyvs_2021}. The poll results are saved as joint frequencies of how many individuals for each possible attribute combination voted for candidate A or B. The polling results can then be used by anomaly detection algorithms as priors or as a part of model training.

\subsection{Fraud Injection}

The types of fraud injection we simulated are vote deletion, vote addition, and vote switching. All forms of fraud occur once the simulation has generated a population, after polling, and before the final results are tallied. Vote deletion represents fraud, where ballots in favor of a certain candidate are thrown out. Vote addition is when additional votes are added to the count in favor of one of the candidates. This type of fraud mimics allegations of fraud where someone who is deceased or didn't vote has their vote cast for them. Vote switching is when the votes of certain individuals are switched over from one candidate to another. More specifically, a certain percentage of votes from a select number of electoral regions are sampled, and if their votes are not cast for a certain candidate, their vote is swapped over in favor of this candidate.

To inject fraud into an election, a number of desired electoral regions to be fraudulent, along with a probability of fraud occurring, needs to be provided. Fraud occurs after the polling occurs and, as of right now, is impartial to the attributes of the individuals it acts upon. Regardless of the mode of voting or attributes like party affiliation, fraud will be equally likely for all individuals in electoral regions with fraud. Limiting fraud injection to two parameters simplifies the process of generating fraudulent data, but the results are not homogeneous because of the differing population distributions of electoral regions. We assume fraud also only occurs in favor of one candidate, which means the fraud impacts groups who vote against the candidate favored by the fraud. Table 2 highlights three sample simulations with the same parameters for their fraud injection.

The injection of fraud alongside the election generation was developed independently from the anomaly detection algorithm we test in order to keep this study unbiased. The levels of fraud and the method to generate the fraud were not known by the research members working on the algorithm. The full labels and parameters of the datasets given to the developers working on the algorithms were not known until after development and testing were complete.

\begin{table}[h!]
\centering
\begin{tabular}{llllllll}
              \midrule
              ER & Poll Pred. & ER with Fraud & Sig. of Fraud  \\
              \midrule
             0.499 & 0.471 & 0.511 & 7/10 \\
             0.515 & 0.509 & 0.525 & 5/10 \\
             0.502 & 0.533 & 0.510 & 4/10 \\
\end{tabular}
\caption{Three independently simulated elections' results. ER refers to the Election Result, the overall popular vote for Candidate A. Poll Pred. refers to the Poll Prediction, the poll result on 5\% of the population. ER with Fraud refers to the Election Results with Fraud, the Election Results with swapping fraud on 1\% of the population (10 affected electoral regions at 25\% population) in favor of Candidate A. Sig. of Fraud refers to the Significance of Fraud, the proportion of fraud infected electoral regions with their majority vote flipping from Candidate B to A due to fraud.}
\end{table}

\subsection{Primary Clustering}
The goal of the primary clustering step is to sort electoral regions into clusters in a way that attempts to maximize demographic similarities between regions in the same cluster. For this part of the formulation, we draw upon \cite{10.1145/3530190.3534799} as they utilize a similar first step.

We first choose a set of demographic variables that are the most critical for the prediction of election results such that they can be clustered upon. To accomplish this, we set the election variables $y$, an E dimensional vector where E is the number of election variables, equal to a weighted sum between $X$ an E x M dimensional matrix where M is the number of variables, the matrix of variables and $\beta$, the E x M+1 dimensional matrix of weights. We then use the AIC score to narrow down the number of predictive variables used.

\begin{align}
    \hat{y} = \beta_0 + \beta_1 x_1 + \beta_2 x_2 + \beta_3 x_3 + ... + \beta_k x_M
\end{align}

Following the selection of variables, we then sort such that intra-cluster deviation in electoral features is minimized. As described in \cite{10.1145/3530190.3534799}, this problem can be summarized as: 

\begin{align}
    \mathop{argmin}_{\textbf{A}} \sum_{i=1}^{k}\sum_{x \in A_{i}}||x-\mu_{i}||^2
\end{align}

where k clusters are formed, $A_i$ represents the regions in each cluster, and $\mu_i$ is the centroid of each cluster. We choose the k-means algorithm to solve this problem because it guarantees convergence, places all regions into clusters, and provides a degree of flexibility in being able to tune k.

\subsection{Novelty Detection}
For the next step, we need to utilize both the polling data and the regression model that was formed for the purposes of variable selection. In terms of the poll, we are provided a number of voters from each covariate demographic that are polled to vote for each candidate. From there, we can use the demographic information from each electoral region to extrapolate how the poll would perform there. In terms of the demographics provided at the electoral region level, we have to use the independence property to estimate the percentage of people from certain covariate demographics. For example, we would use the percentage of white and rich people to estimate the percentage of rich, white people. When there are two candidates and two mutually exclusive classes of variable, i.e. race and wealth, the formulation becomes as follows for each electoral region:

\begin{align}
    \hat{z} = \sum_{d=1}^{D}\sum_{f=1}^{F}
    \alpha_{d}\alpha_{f}(\frac{a_{df}}{a_{df}+b_{df}})
\end{align}

where z is the E dimensional vector of poll results, $\alpha$ describes the percentage of people in each demographic, and a and b denote the number of votes received by candidates a and b, respectively. D and F refer to the sizes of the two classes of mutually exclusive variables.
 
A vector composed of the regression predictions and polling election result predictions, or $\begin{bmatrix}
           \hat{y_i} & \hat{z_i}
            \\
         \end{bmatrix}^T$
         is then entered as input training data into a separate One-class Support Vector Machine for each cluster where the algorithm is described in \cite{scholkopf_platt_shawe-taylor_smola_williamson_2001} and uses a non-Linear RBF kernel. We choose this algorithm as it allows for an input of clean data and tests if new data is similar to the input which is what we are attempting to do. The SVM is parameterized by nu, which describes the probability of detecting an observation that is actually regular outside the decision boundary.

New observations for actual election results are then input into the SVMs for the corresponding clusters to which they belong. The abnormal regions are received as output.

\section{Results}
We apply the model described in 1.3 and 1.4 to the simulation data generated from the methodology described in 1.1 and 1.2. 250 regions with an average of 2000 people per electoral region were generated along with their corresponding demographic data, and polls were created. We first apply the model to a clean dataset without any fraud injected. The model detected 2 fraudulent regions from the 250 clean electoral regions for an accuracy of 99.2$\%$. An example of a cluster and its corresponding SVM can be seen in Figure 2 (a). 

We then test the model on datasets with varying levels of fraud potential applied to a various number of electoral regions. Table 3 displays these results, which include the number of total errors the model made as well as the percent of fraudulent regions that were caught. As you can see from Table 3, precision (fraction of regions identified by the model to be fraudulent that are actually fraudulent), recall (percent of fraudulent regions that the model caught to be fraudulent), and accuracy are calculated and show increased performance when higher levels of fraud are present. We can see that at a 5$\%$ fraud level, the model has no sensitivity to detecting fraud but still has a relatively high accuracy due to the fact that most of the electoral regions are non-fraudulent. It is also important to note the scale at which this fraud is happening. For example at a 5$\%$ fraud level, if 10 regions are affected, 1000 votes are affected by fraud out of a million votes. At higher levels of fraud, the model is able to recover the fraudulent regions well.

\begin{figure}[htb]

\begin{minipage}[b]{\linewidth}
  \centering
  \centerline{\includegraphics[width=7cm]{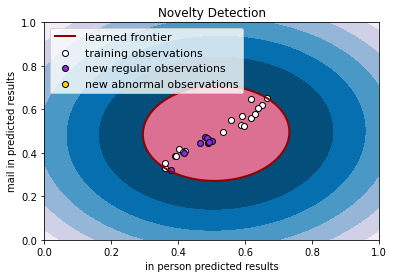}}
%  \vspace{2.0cm}
  \centerline{(a)}\medskip
\end{minipage}

\begin{minipage}[b]{\linewidth}
  \centering
  \centerline{\includegraphics[width=7cm]{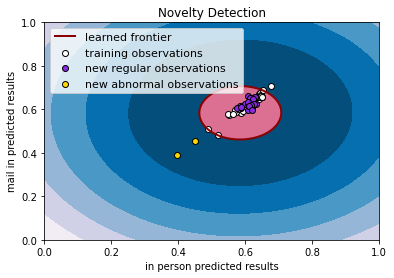}}
%  \vspace{1.5cm}
  \centerline{(b)}\medskip
\end{minipage}

\label{fig:southshore_stat_cali}
\caption{(a) The SVM for a cluster within the dataset with no injected fraud (b) The SVM for a cluster within the dataset with 40 electoral regions that have 40 percent fraud potential.}
\end{figure}

\begin{table*}[h!]

\centering
\begin{tabular}{llllllll}
                \midrule
              Fraud Level* & 4$\%$FR**  & 10$\%$FR        & 16$\%$FR &  precision &  recall &  accuracy     \\
              \midrule
             5 &         .4/0 & 4.8/.4    & 4.8/1.2 &16&5.33 &90.53 \\
  12.5 &        8.8/3.2  & 11.6/6.4     & 10.4/9.6 &62.33&64 &92.53      \\
             20        & 5.6/3.2  & 10.8/10     & 17.6/14.4&81.17&92 &97.07  \\
\end{tabular}
\caption{Fraud detection results (all results in $\%$) -*Fraud level is defined as the estimated percentage of votes actually affected by fraud and is the result of a higher number of votes being subject to potential fraud.** $\%$FR refers to percent of fraudulent regions being affected by the fraud level and the results are shown in the form of  total errors detected / true fraudulent regions detected.  }
\end{table*}

\section{Baseline Comparison}
We compare the results of our methodology to two published state-of-the-art models (\cite{10.1145/3530190.3534799} and \cite{zhang_alvarez_levin_2019}) as baselines. We will refer to \cite{10.1145/3530190.3534799} as Baseline 1 and \cite{zhang_alvarez_levin_2019} as Baseline 2. Baseline 1 differs from our model in that while it also partitions on demographics before attempting to detect anomalies, it is implemented in an unsupervised fashion and does not take into account polling data. 

Baseline 2 uses a completely different approach to simulating clean electoral regions using data from past elections and polling and then injecting fraud into some regions. That data is used to train a Random Forest model which is then applied to real-world data. For the purposes of our baseline, we use a simulation with 125 fraudulent regions (with an equal mix of 5,12.5, and 20 percent fraud levels) and 125 not fraudulent regions to train the model and then apply the model to the test sets of interest. Both the training and test sets utilize the same Neural Network, and all regions are additionally affected by random variance. In the regular case, the training simulation would be built with data from past elections, but currently our simulation does not support simulations with temporal aspects; however, this is a feature we are working on. Thus, in actuality, the training set would likely have contained more noise than the training set that we use. As both the training simulation and test simulation in this case essentially originate from the same ground truth with the difference of random variance added to each region, we expect this baseline to perhaps perform better in this comparison than in actuality where it does not have access to such data. One noted criticism of this method is that data from old elections, which is used to simulate a clean election, likely also contains fraud if the election in question contains fraud. Thus, it would not be truly clean. That would not, however, be an issue here. 

\begin{table*}[h!]

\centering
\begin{tabular}{llllllll}
            \midrule
              Fraud Level & 4$\%$FR  & 10$\%$FR        & 16$\%$FR &  precision &  recall &  accuracy     \\
              \midrule
             5 &         6.4/.4 & 7.6/.8    & 10.4/1.6 &11.48&9.33 &83.73 \\
  12.5 &        5.6/0  & 4.4/2     & 7.6/3.6 &31.82&18.60 &87.87      \\
             20        & 5.6/2  & 7.6/3.6     & 12.8/5.2&41.54&36.00 &88.53  \\
\end{tabular}
\caption{Fraud detection results for Baseline 1 (all results in $\%$)}
\end{table*}

We can see from Table 4 that while Baseline 1 detects close to the correct amount of fraud in most situations, the regions that it identifies to be fraudulent are usually not correct leading to low precision. Baseline 1 also fails to catch many of the actually fraudulent regions leading to the low recall. This could possibly be attributed to the methodology wherein the authors use DBSCAN in their anomaly detection step, and if there is enough fraud, the fraudulent regions could be grouped in with other fraudulent regions and therefore not correctly identified by the classifier as anomalous from any cluster. Our methodology generally retains higher precision, recall, and accuracy compared to this baseline.

\begin{table*}[h!]

\centering
\begin{tabular}{llllllll}
\midrule
              Fraud Level & 4$\%$FR  & 10$\%$FR        & 16$\%$FR &  precision &  recall &  accuracy     \\
              \midrule
             5 &         39.2/2.4 & 39.6/8    & 38.8/10.8 
             &18.03&	70.67&	64.93
\\
  12.5 &        14.4/2  & 22.8/8.4     & 27.6/14.4 &
  38.27&	82.67&	84.93
\\
             20        & 10.8/4  & 22.4/10     & 24.8/16 &
             51.72&	100&	90.67
\\
\end{tabular}
\caption{Fraud detection results for Baseline 2 (all results in $\%$)}
\end{table*}

As we can see from Table 5, Baseline 2 has a high recall, but low precision. This means it is highly prone to the detection of false positives. As this baseline generally has lower precision and higher recall than our model, we compare the models by using the F1 score which combines the two metrics into one by using their harmonic mean. 

\begin{table}[h!]
% \toprule
\centering
\begin{tabular}{llllllll}
              Fraud Level & F1 PM*  & F1 B2**& Acc. PM& Acc. B2  \tabularnewline
              \midrule
             5&4.00&14.37&90.53&64.93 \tabularnewline
              12.5&31.58&  26.16&92.53&84.93 \tabularnewline
              20& 43.12& 34.09&97.07&90.67 \tabularnewline
              Average & 26.23&24.87&93.38   &86.71
\end{tabular}
\caption{Comparison of F1 Scores and Accuracy - *PM refers to our proposed model, **B2 refers to Baseline 2 }
\end{table}

We can see from Table 6 that our proposed model has a higher average F1 score than Baseline 2. However, Baseline 2 has a far better recall at the 5 percent fraud level case leading to it having a higher F1 score at that level. Our model retains a higher accuracy at every level, including a significantly higher accuracy at the 5 percent fraud level. Based on these results, we would argue that both Baseline 2 and our model should not be used for the detection of extremely low amounts of fraud. Our model generally fails to catch these cases while Baseline 2 does catch these cases but identifies nearly half of all regions to be fraudulent and as such has low enough precision to make its use infeasible. At higher levels of fraud, both metrics F1 and accuracy favor our model. However, on a more specific level, the usage of models does depend on the preferences of the entity utilizing the model. For example, if it is more important to the user to catch almost all fraudulent cases and if the user is fine with having more false positives, Baseline 2 may be preferred. If the user wants to penalize false positives while being able to still catch most of the fraudulent cases but likely less than in Baseline 2, our model may be preferred.

\section{Discussion}
The simulation was able to successfully generate clean and fraudulent datasets that were used to evaluate model performance. An application that we explored in this paper would be to compare the performance of different anomaly detection methodologies in detecting fraud with various setups. The simulation has the advantage of being able to control what specific types of fraud and how much of said fraud is going to be injected. By evaluating various anomaly detection methodologies on different types of fraud, one could determine which methodologies best detect different types and amounts of fraud. This could help with fine-tuning the process of election fraud detection. If there is an election that has many allegations of deletion fraud, a methodology that was proven to perform best in detecting deletion fraud in simulated data could be used to detect fraud in this scenario. To expand upon this, the simulation could be expanded to have a temporal aspect as well. This way we could generate data for elections taken over time as populations shift and voting patterns change. Older election data could be treated as prior election results which could be used by algorithms to detect anomalous changes in an electoral region's voting over time. Another direction to take the simulation would be generalizing it so that it can model elections with more than two candidates. 

 The model is highly applicable to a variety of situations because it requires only data that is generally publicly available as well as the fact that is holistic and is not designed to only catch one type of fraud. Our research adds to the existing body of work by introducing a supervised detection technique as well as introducing polling as a potential source of supervision. Although our model performs better than others in this regard, one limitation remains the significant number of false positives detected even at high fraud levels. As such, this model is best not used to definitively declare fraud, but instead to suggest possibly fraudulent regions and to narrow down resource allocation. Another limitation is the quality of the data itself. In our simulation, we ran the polls with an appropriately small margin of error, but election polls will not always have consistently good results. If the polls are done badly for some reason, the method will likely not perform as well. In future work, we plan to conduct a more extensive analysis of how sensitive the model is to errors in polling. Additionally, a potential future direction for the model could be to incorporate temporal data into the novelty detection step.

% Use \bibliography{yourbibfile} instead or the References section will not appear in your paper
\bibliography{aaai23}

\end{document}